\documentclass{article}
\PassOptionsToPackage{numbers}{natbib}
\usepackage[final,dblblindworkshop]{neurips_2026}
\workshoptitle{AI for Chip Design}
\makeatletter
\renewcommand{\@noticestring}{Accepted at the AI for Chip Design Workshop,
  \@neuripsordinal\ Conference on Neural Information Processing Systems
  (NeurIPS \@neuripsyear).}
\makeatother
\usepackage{booktabs,graphicx,url,amsmath,xcolor}
\usepackage{tikz}
\usetikzlibrary{positioning,arrows.meta,fit,calc}
\definecolor{linkblue}{RGB}{40,96,150}
\usepackage[colorlinks=true,allcolors=linkblue]{hyperref}
\title{SGAnalog: An End-to-End Circuit Benchmark\\
from Open-Source Silicon Tapeouts}
\author{%
  Yueting Li \\
  University of California, Berkeley \\
  \texttt{yueting\_li@berkeley.edu} \\
  \And
  Weihang Ding \\
  University of California, Berkeley \\
  \texttt{dingcharles@berkeley.edu} \\
}

\begin{document}
\maketitle

\begin{abstract}
Existing analog integrated circuit design benchmarks make two questions hard
to answer: whether a model has learned transferable circuit skills rather than
recalled familiar examples, and whether its output works under defined process
and test conditions. We introduce a benchmark built from human-designed,
open-source circuits associated with Tiny Tapeout manufacturing shuttles.
The collection contains 273 topologically distinct top-level designs.

Every source is retrieved at the revision recorded for its shuttle submission
and processed in a fixed containerized environment. The pipeline exports each
eligible schematic image and its SPICE netlist from the same source file,
giving transcription an exact structural reference. Commit dates support
model-specific training-cutoff analysis, while author testbenches provide the
simulation context for sizing.

The benchmark evaluates schematic-to-netlist transcription and device sizing.
Across seven models on a fixed set of 66 transcription tasks, the strongest model
reaches 56.1\% exact graph isomorphism, and six of seven models drop sharply from
the small to the medium tier. For one frontier model, removing author-chosen labels
reduces exact matches while preserving aggregate structural F1, suggesting that
labels can aid connectivity tracing. On the 17 sizing tasks, the leading model
converges on all 17 proposals and reaches 91.2 out of 100 against the human
reference, while the two newest Claude models refuse 4 and 11 of the same
prompts they transcribe without objection; a proposal without sizes scores
zero. The two tasks produce different model rankings, exposing
distinct visual and design capabilities and, in one family, a policy rather
than capability limit.
\end{abstract}

\section{Introduction}

Analog integrated circuit design gives model outputs a concrete standard:
the topology must be correct, and the sized circuit must work under defined
process and test conditions. Evaluating these capabilities requires more than
a collection of familiar circuit examples. The benchmark must also expose
possible training overlap and provide an executable reference for function.
Existing resources rarely provide both.

\begin{figure}[t]
\centering
\definecolor{flowB}{RGB}{40,96,150}\definecolor{flowF}{RGB}{228,240,250}
\definecolor{dataF}{RGB}{204,226,245}
\definecolor{tealB}{RGB}{20,110,100}\definecolor{tealF}{RGB}{222,242,238}
\definecolor{orgB}{RGB}{178,98,20}\definecolor{orgF}{RGB}{253,238,219}
\definecolor{redB}{RGB}{168,56,46}\definecolor{redF}{RGB}{252,233,230}
\definecolor{dockB}{RGB}{29,99,180}\definecolor{dockF}{RGB}{226,238,250}
\begin{tikzpicture}[
  inbox/.style={draw=flowB, fill=flowF, rounded corners=2pt, align=center,
              font=\scriptsize, inner sep=3.5pt, minimum height=11mm},
  frame/.style={draw, rounded corners=3pt, inner sep=4.5pt, semithick},
  ftitle/.style={font=\scriptsize\bfseries, fill=white, inner sep=2.5pt},
  arr/.style={-{Stealth[length=2mm]}, semithick, draw=flowB},
  alab/.style={font=\tiny, text=black!55, fill=white, inner sep=1pt},
  garr/.style={densely dashed, draw=redB}]

\node[inbox] (tt) {\textbf{Tiny Tapeout}\\27 shuttles\\2023--2026};
\node[inbox, right=4.5mm of tt] (src) {\textbf{Design sources}\\schematics, testbenches\\at the tapeout revision};
\node[inbox, fill=dataF, thick, right=4.5mm of src] (data)
  {\textbf{The benchmark}\\schematic image $\leftrightarrow$ netlist\\$+$ author testbench};
\draw[arr] (tt) -- (src);
\draw[arr] (src) -- (data);
\node[frame, draw=flowB, fit=(tt)(src)(data)] (dsF) {};
\node[ftitle, text=flowB, anchor=west] at ($(dsF.north west)+(2.5mm,0)$)
  {Dataset construction};

\node[inner sep=1pt, right=19.5mm of data, yshift=-0.5mm] (tools)
  {\begin{tabular}{@{}c@{}}
   \raisebox{-1.5mm}{\includegraphics[width=4.2mm]{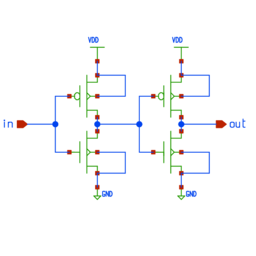}}\,{\scriptsize Xschem}\hspace{2mm}\includegraphics[width=10.5mm]{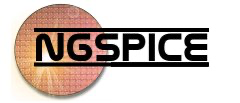}\\[1.7mm]
   \includegraphics[width=13mm]{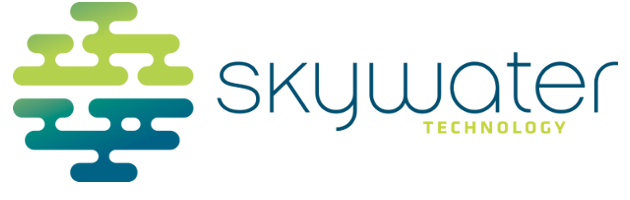}
  \end{tabular}};
\node[frame, draw=dockB, fit=(tools), inner sep=3.5pt] (dockF) {};
\node[anchor=west, fill=white, inner sep=1.5pt] (docklogo)
  at ($(dockF.north west)+(1.8mm,0)$)
  {\includegraphics[width=14mm]{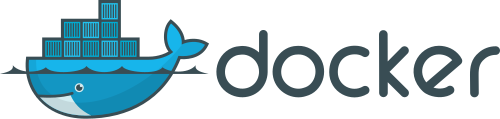}};
\node[below=2.2mm of dockF, align=center, font=\scriptsize] (iface)
  {\textbf{Benchmark interface}\\evaluation harness, simulation browser};
\node[frame, draw=black!45, fit=(dockF)(docklogo)(iface)] (envF) {};
\node[ftitle, text=black!60, anchor=west] at ($(envF.north west)+(2.5mm,0)$)
  {Environment};

\coordinate (mid) at ($(dsF.south)!0.5!(dsF.south -| envF)$);
\node[inbox, draw=tealB, fill=tealF, thick, below=19mm of mid,
      anchor=north east, xshift=-2.5mm]
  (t1) {\textbf{Task 1: read the schematic}\\schematic image $\rightarrow$ SPICE netlist\\score: netlist connectivity matches the human design\\[9.5mm]\mbox{}};
\node[inbox, draw=orgB, fill=orgF, thick, below=19mm of mid,
      anchor=north west, xshift=2.5mm]
  (t2) {\textbf{Task 2: size the devices}\\unsized topology $+$ testbench $\rightarrow$ W/L\\score: simulated output ratio to the human design\\[9.5mm]\mbox{}};
\node[frame, draw=black!45, fit=(t1)(t2)] (tkF) {};
\node[ftitle, text=black!60, anchor=west] at ($(tkF.north west)+(2.5mm,0)$)
  {Tasks};

\begin{scope}[shift={($(t1.south)+(-1mm,6.8mm)$)}, line cap=round,
              every path/.style={draw=tealB, semithick}]
\draw (-22mm,2mm) -- (-22mm,-1mm);
\draw (-22mm,2mm) -- (-19.6mm,2mm);
\draw (-19.2mm,2mm) circle (0.4mm);
\draw (-18.8mm,0.9mm) -- (-18.8mm,3.1mm);
\draw (-18.2mm,0.7mm) -- (-18.2mm,3.3mm);
\draw (-18.2mm,2.8mm) -| (-16.6mm,4.2mm);
\draw (-18.2mm,1.2mm) -| (-16.6mm,0.5mm);
\draw (-22mm,-1mm) -- (-18.8mm,-1mm);
\draw (-18.8mm,-2.1mm) -- (-18.8mm,0.1mm);
\draw (-18.2mm,-2.3mm) -- (-18.2mm,0.3mm);
\draw (-18.2mm,-0.2mm) -| (-16.6mm,0.5mm);
\draw (-18.2mm,-1.8mm) -| (-16.6mm,-3.2mm);
\draw (-16.6mm,4.2mm) -- (-16.6mm,4.9mm);
\draw (-16.6mm,-3.2mm) -- (-16.6mm,-3.9mm);
\draw (-16.6mm,0.5mm) -- (-14.6mm,0.5mm);
\draw[-{Stealth[length=1.6mm]}] (-12.6mm,0.5mm) -- (-9.6mm,0.5mm);
\end{scope}
\node[font=\tiny\ttfamily, align=left, anchor=west, text=tealB!60!black]
  at ($(t1.south)+(-9.6mm,7.3mm)$)
  {XM1 out in VDD pfet\\XM2 out in VSS nfet};

\begin{scope}[shift={($(t2.south)+(-6mm,6.6mm)$)}, line cap=round,
              every path/.style={draw=orgB, semithick}]
\draw (-14.4mm,0.7mm) -- (-11.9mm,0.7mm);
\draw (-11.9mm,-0.4mm) -- (-11.9mm,1.8mm);
\draw (-11.3mm,-0.6mm) -- (-11.3mm,2mm);
\draw (-11.3mm,1.5mm) -| (-9.7mm,2.9mm);
\draw (-11.3mm,-0.1mm) -| (-9.7mm,-1.5mm);
\draw[-{Stealth[length=1.6mm]}] (-6.5mm,0.7mm) -- (-3.5mm,0.7mm);
\draw (0mm,0.7mm) sin (2mm,3mm) cos (4mm,0.7mm) sin (6mm,-1.6mm)
      cos (8mm,0.7mm) sin (10mm,3mm) cos (12mm,0.7mm);
\draw[draw=orgB!45] (0mm,0.7mm) sin (2mm,2mm) cos (4mm,0.7mm)
      sin (6mm,-0.6mm) cos (8mm,0.7mm) sin (10mm,2mm) cos (12mm,0.7mm);
\end{scope}
\node[font=\tiny, text=orgB!70!black, anchor=north]
  at ($(t2.south)+(-17.8mm,5.6mm)$) {W/L\,=\,?};

\draw[arr] (dsF.south -| data) --
  node[alab, fill=none, pos=0.5, anchor=east, xshift=-1.5mm]
  {tasks draw on the benchmark}
  (tkF.north -| data);
\draw[arr, draw=dockB] (envF.west |- data) --
  node[alab, fill=none, pos=0.5, anchor=south, yshift=0.6mm] {builds}
  (dsF.east |- data);
\draw[arr, draw=dockB] (envF.south) --
  node[alab, fill=none, pos=0.55, anchor=west, xshift=1.5mm]
  {simulates, grades}
  (tkF.north -| envF.south);

\node[frame, draw=redB, fill=redF, densely dashed,
      below=7.5mm of tkF.south, anchor=north,
      text width=129mm, align=center, font=\scriptsize, inner sep=4pt]
  (guard) {\textbf{Memorization defenses}:
  commit dates vs.\ training cutoffs, \mbox{label-stripped schematics},
  canary in every netlist, no tools at evaluation,
  new shuttles every year};
\draw[garr] (guard.north -| t1) -- (tkF.south -| t1);
\draw[garr] (guard.north -| t2) -- (tkF.south -| t2);
\end{tikzpicture}
\caption{Benchmark overview. Versioned Tiny Tapeout sources are converted into
matched schematic images and netlists, with author testbenches retained for
simulation. Transcription is scored by device--net graph isomorphism, the
comparison at the core of an LVS check; sizing is scored by simulation;
and potential exposure is analyzed with commit-aware controls.}
\label{fig:overview}
\end{figure}

\paragraph{Separating capability from recall.}
Many analog benchmarks draw from canonical textbooks~\cite{razavi,graymeyer},
published circuit figures~\cite{analoggym,analoggenie}, or synthetic
enumeration~\cite{ocb}. These sources are useful, but their public availability
makes prior model exposure difficult to rule out. Recent systems already
approach saturation on the solved-task count of
AnalogCoder~\cite{analogcoder,analogcoderpro}. Without dated provenance and an exposure
protocol, a high score does not reveal whether the model generalized or
recalled a familiar design.

\paragraph{Verifying function.}
Large datasets often provide topology without device sizes, testbenches, or a
process design kit (PDK). Simulator-backed benchmarks provide the missing
context but cover a small set of base circuits and expand it through variants.
The field therefore lacks a broad set of human designs that supports both
structural comparison and execution under the original test conditions.

\paragraph{Why open-source tapeouts.}
Tiny Tapeout~\cite{tt} provides a different foundation. Its machine-readable
index links each shuttle submission to an exact repository revision. The
analog repositories collectively contain transistor-level schematics,
netlists, PDK context, and author testbenches. The recorded revisions make the
pipeline reproducible, shuttle dates support model-specific cutoff analysis,
and new shuttles provide a continuing refresh path. Verified open-source
licenses allow the resulting artifacts to be redistributed. The collection
spans amplifiers and comparators; ADC, TDC, and DAC blocks; oscillators and
timing circuits; power, bias, and reference circuits; RF front ends and
filters; and sensing interfaces. It also includes smaller passive, switching,
and memory-related subcircuits. The fixed evaluation sets are
eligibility-filtered subsets of this broader collection rather than
equal-sized samples of every family.

\paragraph{Scope.}
The benchmark targets schematic-level reasoning, not layout or subjective
design style. Transcription measures whether a model can recover circuit
structure from an image. Sizing measures whether model-selected device
dimensions preserve simulated behavior in the author's testbench. Topology
generation remains future work because it requires a uniform design brief and
automatic testbench alignment.

\paragraph{Contributions.} We call the benchmark SGAnalog, for
silicon-grade analog.\footnote{Code and data:
\url{https://github.com/Li-Yueting/SGAnalog}.}
\begin{enumerate}
  \item A license-verified, commit-pinned benchmark from open-source Tiny
        Tapeout submissions, built with a deterministic containerized pipeline,
        graph-hash grouping, and aligned schematic images and netlists
        generated from the same source (\S\ref{sec:construction}).
  \item Two end-to-end evaluations: graph-scored schematic transcription and
        simulator-scored device sizing against the human reference under the
        same PDK and author testbench (\S\ref{sec:tasks}).
  \item An exposure-aware protocol combining commit dates with model-specific
        cutoffs, closed-book evaluation, blinded renders, release canaries, and
        refreshes from new shuttles (\S\ref{sec:contamination}).
  \item A seven-model baseline showing a broad small-to-medium transcription
        drop and different rankings across transcription and sizing
        (\S\ref{sec:tasks}).
\end{enumerate}

\section{Related work}\label{sec:related}

\begin{table}[h]
\centering\small
\setlength{\tabcolsep}{3pt}
\caption{Analog benchmark landscape. Sized = devices carry numeric sizes.
TB = author testbenches are included. PDK = an open process design kit
simulation flow is included. ``part''\ marks partial coverage. Audit = the
benchmark ships a contamination audit. Scale follows
the unit reported by each source and is therefore not directly comparable.}
\label{tab:landscape}
\begin{tabular}{@{}lllllll@{}}
\toprule
Benchmark & Scale & Provenance & Sized & TB & PDK & Audit \\
\midrule
AnalogGenie~\cite{analoggenie}     & 3,350 topo.        & textbooks + papers   & no  & no   & no   & no \\
CktGNN / OCB~\cite{ocb}            & 10k op-amps        & synthetic            & part& part & no   & no \\
AICircuit~\cite{aicircuit}         & 9 topo.\ / 361k pts& authored             & yes & yes  & part & no \\
OSIRIS~\cite{osiris}               & 5 templ.\ / 87.1k layouts & authored      & yes & yes  & yes  & no \\
MAGICAL-CIRCUITS~\cite{magical}    & 26 blocks          & UT Austin silicon    & yes & part & no   & no \\
AnalogCoder~\cite{analogcoder}     & 24 tasks           & authored             & part& yes  & no   & no \\
AMSnet 1.0 / 2.0~\cite{amsnet}     & 734 / 2,686        & textbook scans       & no  & no   & no   & no \\
Masala-CHAI~\cite{masala}          & 7,500              & textbook scans       & no  & no   & no   & no \\
AnalogGym~\cite{analoggym}         & 30 topo.           & published papers     & part& part & part & no \\
\midrule
\textbf{SGAnalog (ours)} & \textbf{273 distinct topologies} & \textbf{Tiny Tapeout repos} &
\textbf{yes} & \textbf{part} & \textbf{yes} & \textbf{yes} \\
\bottomrule
\end{tabular}
\end{table}

Prior image-based sets such as AMSnet~\cite{amsnet} and
Masala-CHAI~\cite{masala} pair scanned textbook figures with netlists produced
by manual or model-assisted extraction. In this benchmark, every eligible
image--netlist pair is generated from one schematic source, avoiding a separate
annotation step.

AnalogGym~\cite{analoggym} is the closest simulator-backed benchmark in
intent. Its amplifier topologies come from published literature, whereas our
sizing reference is the submission author's design evaluated under the same
PDK and testbench as the model. Concurrent work,
NotSoTiny~\cite{notsotiny}, uses the digital side of Tiny Tapeout~\cite{tt} for RTL code
generation; our benchmark targets the analog artifacts and their simulation
context.

\section{Dataset construction}\label{sec:construction}

\begin{table}[t]
\centering\small
\caption{Dataset spine: from the shuttle index to the fixed task sets.
Appendix~\ref{app:inventory} gives the full triage, licensing,
composition, and redundancy inventory.}
\label{tab:funnel}
\begin{tabular}{@{}lr@{}}
\toprule
From index to task sets & $n$ \\
\midrule
Index projects (27 shuttles) & 4,572 \\
Entries with analog pins     & 237 \\
Repositories contributing circuits & 125 \\
Schematics fetched at pinned commits & 1,448 \\
Circuits & 933 \\
Top-level designs & 425 \\
Unique device--net graphs & 273 \\
Fixed tasks: transcription / sizing & 66 / 17 \\
\bottomrule
\end{tabular}
\end{table}

\paragraph{Harvest funnel.} Table~\ref{tab:funnel} walks the index down
to the fixed task sets; Appendix~\ref{app:inventory} holds the full
triage, licensing, and composition inventory. Blobless clones triage every candidate by file tree
and license without downloading design data, licensing is verified from
the license texts as fetched at the pinned commits rather than assumed,
and design source is fetched at the commit the shuttle index records,
never at HEAD.

\paragraph{Circuit identification.} Counting circuits naively fails in both
directions: every \texttt{.sch} overcounts (symbols, testbenches,
sub-blocks); counting only hierarchy tops undercounts, because the standard
xschem~\cite{xschem} idiom places the interesting circuit under its
testbench. We
classify by role instead. A \emph{testbench} contains stimulus and
simulation control, or matches the naming conventions for one. A
\emph{circuit} is any non-testbench schematic. A \emph{top-level design}
is a circuit that no other non-testbench schematic instantiates.
Table~\ref{tab:inventory} (Appendix~\ref{app:inventory}) gives the
resulting composition by role, tier, and era. The trivial
tier also includes transistor-free passive networks and device models such as
photodiode equivalents.

\paragraph{Redundancy.} Repository counts overstate the diversity of the
tapeout-derived collection because blocks are resubmitted and forks preserve
earlier circuits. We compute a strict Weisfeiler--Lehman hash of each top-level
device--net bipartite graph after removing net names. Repeated pad models,
inverters, tutorial circuits, and R-2R blocks account for most shared hashes.
The resulting candidate duplicate groups are released for split control.
Task~1 scoring uses explicit graph isomorphism rather than hash equality.

\paragraph{Reproducibility contract.} Duplicate file stems and archived copies
can make hierarchy resolution depend on filesystem order, so candidates are
resolved by depth and then full path. Every stage, from harvest through
simulation, executes inside one pinned container image
(IIC-OSIC-TOOLS~\cite{iic}, digest \texttt{sha256:ae380e2b\dots df38}),
which pins ngspice~\cite{ngspice}
and supplies the open PDKs; the host requires only Docker. Classification
tables from two complete runs are byte-identical, and simulation
tables regenerate identically up to their per-run timing columns.

\paragraph{Task-set construction.} A curated core spans circuit classes,
difficulty tiers, and eras. For transcription, leaf schematics with every
symbol resolved are eligible; deterministic round-robin sampling over era and
tier fixes 66 tasks. For sizing, we audit 412 author-testbench runs after
deterministic include-path repair, without changing circuits or stimuli. A run
is eligible only if it completes within 600 seconds, contains real device
cards, and resolves the schematic hierarchy; 190 runs meet this gate.
End-to-end device-under-test mapping and the primitive-only hierarchy
requirement then yield 17 tasks. Both sets remain fixed across models.
The release ships a browser interface over the verified
circuit--testbench pairs (Appendix~\ref{app:ui}).
Hierarchical sizing tasks await the measurement layer described in
Section~\ref{sec:roadmap}.

\section{Tasks and evaluation}\label{sec:tasks}

The benchmark defines two complementary tasks. Transcription tests schematic
reading; sizing tests device choices under simulation.

\begin{figure}[t]
\centering
\begin{minipage}[c]{0.55\textwidth}
\includegraphics[width=\textwidth]{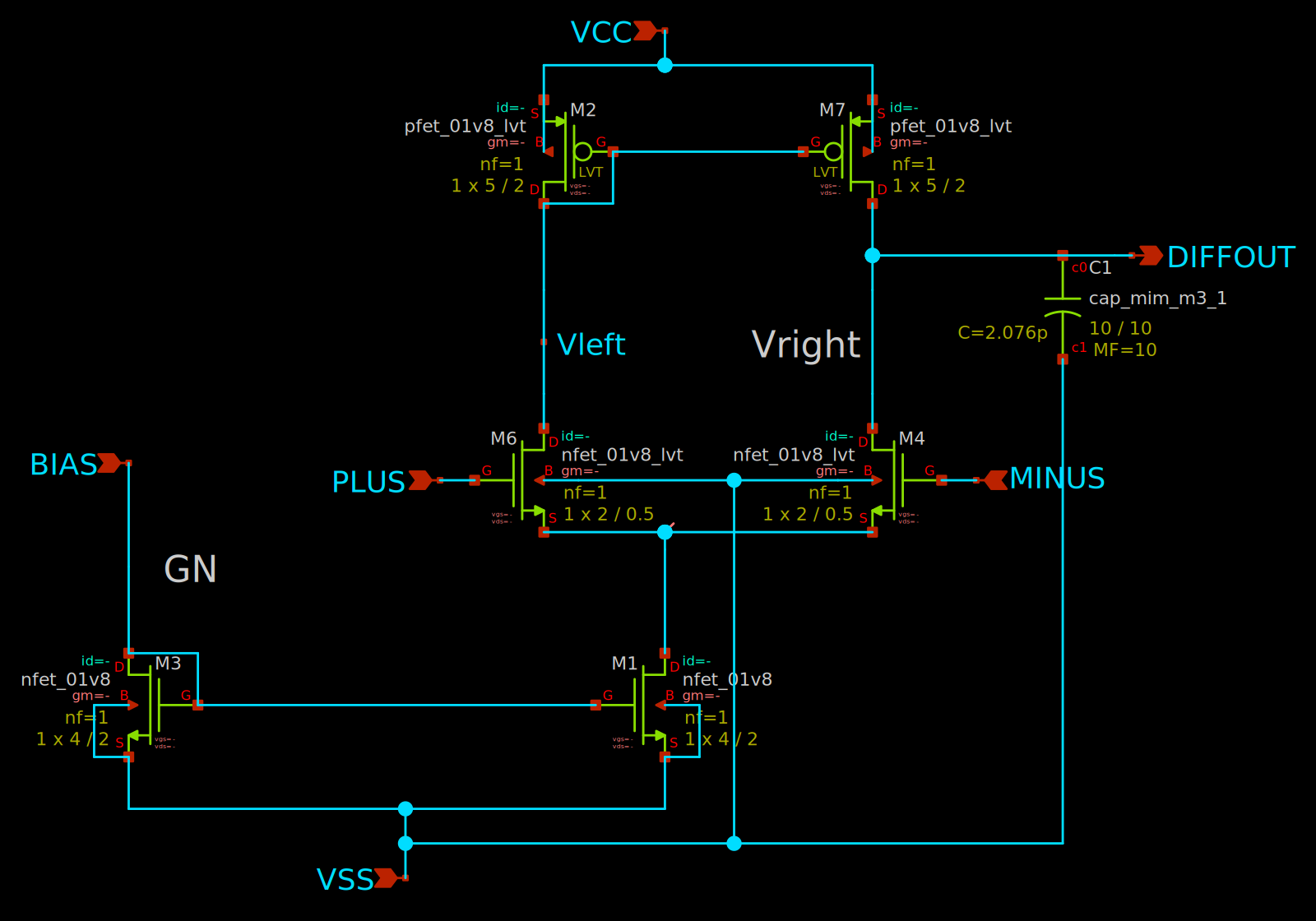}
\end{minipage}\hfill
\begin{minipage}[c]{0.43\textwidth}
\tiny
\begin{verbatim}
* Differential amplifier
* (5T OTA with output cap)

XM2 Vleft Vleft VCC VCC
+ sky130_fd_pr__pfet_01v8_lvt
+ L=2 W=5 nf=1
XM7 DIFFOUT Vleft VCC VCC
+ sky130_fd_pr__pfet_01v8_lvt
+ L=2 W=5 nf=1
XM6 Vleft PLUS TAIL VSS
+ sky130_fd_pr__nfet_01v8_lvt
+ L=0.5 W=2 nf=1
XM4 DIFFOUT MINUS TAIL VSS
+ sky130_fd_pr__nfet_01v8_lvt
+ L=0.5 W=2 nf=1
XM3 BIAS BIAS VSS VSS
+ sky130_fd_pr__nfet_01v8 L=2 W=4 nf=1
XM1 TAIL BIAS VSS VSS
+ sky130_fd_pr__nfet_01v8 L=2 W=4 nf=1
XC1 DIFFOUT VSS
+ sky130_fd_pr__cap_mim_m3_1
+ W=10 L=10 MF=10 C=2.076p
\end{verbatim}
\scriptsize
\textbf{Scored:} graph-isomorphic; W/L accuracy 1.00.
\end{minipage}

\vspace{3mm}
\begin{minipage}[c]{0.55\textwidth}
\includegraphics[width=\textwidth]{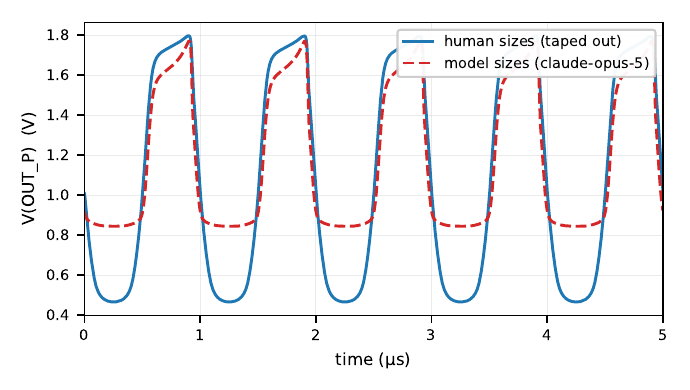}
\end{minipage}\hfill
\begin{minipage}[c]{0.43\textwidth}
\scriptsize
\centering
\begin{tabular}{lll}
\toprule
Device & Human & Model \\
\midrule
input pair      & 3.6/0.38 $\times$2  & 20/0.5 $\times$10 \\
loads (PMOS)    & 3.6/0.38 $\times$2  & 5/0.5 $\times$2 \\
tail            & 3.6/0.38 $\times$2  & 7.2/0.38 $\times$4 \\
bias mirror     & 3.6/0.38 $\times$2  & 3.6/0.38 $\times$2 \\
bias resistor   & L=1.85 $\times$10   & L=30 $\times$1 \\
\bottomrule
\end{tabular}

\vspace{1.5mm}
\raggedright
Sizes as W/L in \textmu m $\times$ nf. In the same
testbench the human sizing swings $\pm$665 mV at
the output; the model's reaches $\pm$463 mV with
a compressed lower rail. The delta is what
simulation-based scoring measures.
\end{minipage}
\caption{Both tasks, end to end. Top: transcription. The input schematic
image (left) and claude-opus-5's verbatim output netlist (right) for a
five-transistor OTA. Its net names differ from the drawing, but its graph is
isomorphic to the reference. Bottom: sizing for a fully differential
amplifier. The model receives the topology with sizes removed, then its proposal
and the human reference run in the same author testbench. Aggregate results
appear in Table~\ref{tab:matrix}.}
\label{fig:pair}
\end{figure}

\paragraph{Task 1: schematic-to-netlist transcription.} Input: the rendered
schematic (Figure~\ref{fig:pair}). Output: a netlist. Scoring is by graph isomorphism, never by text
match. We build a device-net bipartite multigraph from each netlist and test
it for isomorphism against the ground truth, treating the drain and source
of a MOSFET as interchangeable and ignoring net names. We report exact
isomorphism, a structural F1 over device and net signatures for partial
credit, and W/L parameter accuracy. Because image and netlist derive from the
same \texttt{.sch} file, each image corresponds exactly to its reference
netlist, and render style becomes a controllable nuisance variable.

Transcription is an established reading task rather than a design task;
AMSnet~\cite{amsnet} and Masala-CHAI~\cite{masala} study the same extraction
problem. Here it also provides a common evaluation across circuits regardless
of testbench state, enabling the commit-date analysis below.

The baseline below uses the default render. The release also ships a blinded
render for each schematic. Device terminals, size annotations, and PDK model
names remain because reading them is part of the task. Universal supply names
also remain because they carry little circuit-specific retrieval value.
Author-chosen net labels and instance names are removed as plausible retrieval
handles; removal is driven by each circuit's netlist rather than by pattern
matching. The blinded task stays well posed because the scorer ignores names.

On claude-opus-5, blinding lowers exact isomorphism from 51.5\% to 36.4\%
while structural F1 is unchanged and parameter accuracy improves. Because the
scorer ignores names, the result is consistent with labels serving as visual
anchors for connectivity rather than as part of the answer itself. It does not
rule out prior exposure; the cutoff analysis addresses that separate channel.

\paragraph{Task 2: device sizing.} Input: the human topology with W/L
stripped to placeholders, the testbench, the PDK, and target metrics. Output:
sizes; the topology is fixed. The reference is the simulated performance of the
human-authored sizing under the same PDK, testbench, and simulator.
Figure~\ref{fig:pair} (bottom) shows the task
executed end to end on one circuit: the model's proposed sizes beside the
reference sizes, and both simulated in the author's unmodified testbench.

\begin{figure}[t]
\centering
\includegraphics[width=\textwidth]{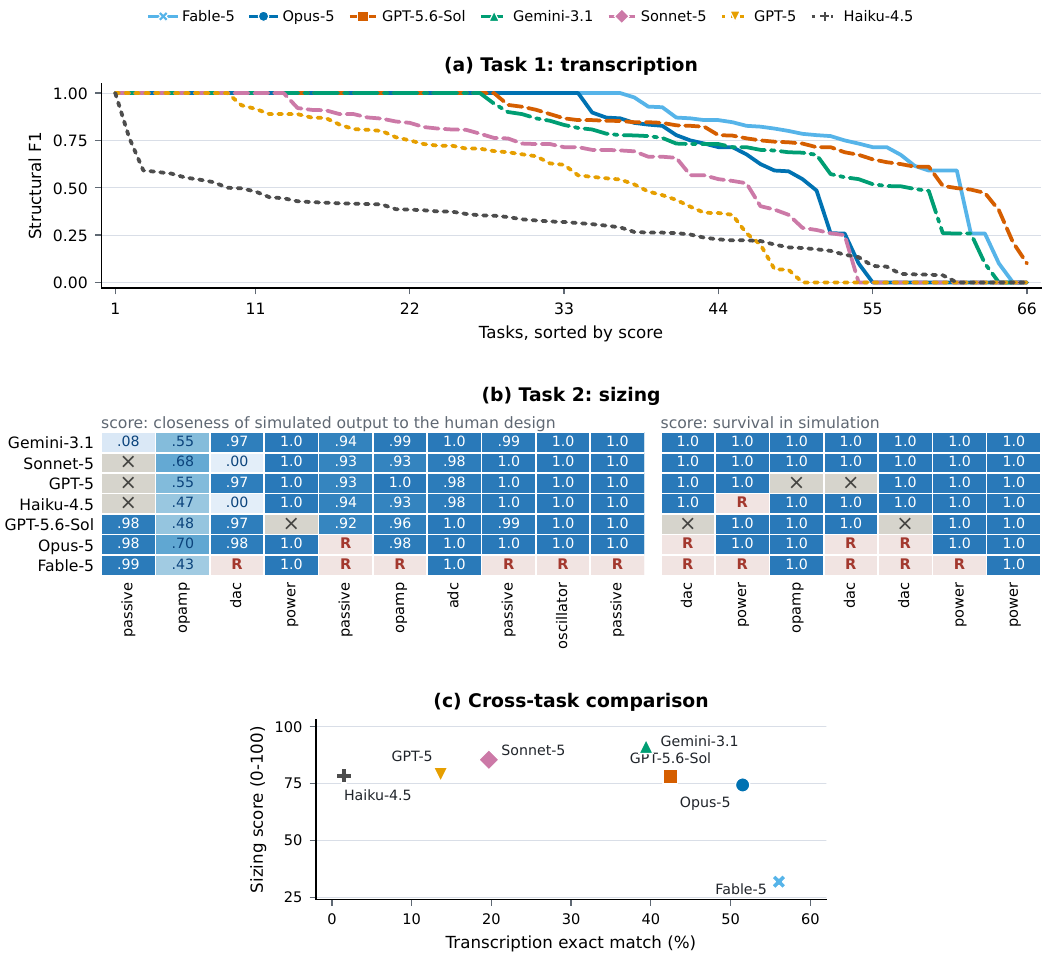}
\caption{Main results for both tasks. (a) Task 1: each model's
66 structural F1 scores sorted in descending order. The width of the plateau
at 1.0 equals the exact-match count, and the tail shape is the failure mode:
Opus falls steeply once it fails, while Sol declines gradually. The encoding
uses no banding thresholds. (b) Task 2: the complete score matrix, seven models
by 17 sizing tasks, one cell per proposal. Each cell prints its score, with
blue depth as a redundant encoding; a gray cross marks a nonconvergent
sizing, and a red R marks a proposal that provided no sizes, which scores
zero. The left group is scored by how close the proposal's simulated
output comes to the human design's, hardest answered task first; in the right
group the author analyses expose no usable scalar, so a convergent sizing
scores one. The two .00 cells,
on a DAC task, are convergent sizings whose output collapsed. (c) Transcription
accuracy versus sizing score: model rankings differ across the two
capabilities. The sizing axis is truncated to keep the seven points legible.
The difficulty-tier breakdown of Task 1 is in Appendix~\ref{app:tiers}.}
\label{fig:curves}
\end{figure}

The task runs on the fixed sizing set (\S\ref{sec:construction}); scoring is
author-testbench-relative. Both sizings run the author's analyses. The
reference selects the output vector with the largest transient swing or
absolute low-frequency AC gain, and the model is evaluated on that same vector.
For each eligible metric, let $r$ be the model-to-reference ratio; the metric
score is $\min(r,1/r)$, averaged across available metrics. A nonconvergent run
scores zero, while a convergent task without an eligible scalar metric receives
a survival score of one. A proposal that provides no sizes also scores zero:
the unmodified deck holds the human sizes, so simulating it would grade the
reference against itself. Table~\ref{tab:matrix} reports
the mean scores, and Figure~\ref{fig:curves}b shows every cell behind them.
Every Claude or Gemini proposal that provides sizes converges,
whereas both GPT generations produce several invalid device geometries.
Convergence alone is too weak a bar: on one task, two proposals converge
with output swing below one percent of the reference, a failure the
convergence count cannot see and the simulated metric does. Refusal is a
third failure mode, and it lands on the newest models: claude-opus-5
returns an empty API-level refusal on 4 of the 17 prompts and
claude-fable-5 on 11, while both transcribe the same
circuits without objection. A full repeat run of claude-fable-5 refuses
12, ten prompts refused both times and three changing sides, so the
refusal set is stable in bulk but not per prompt. On the tasks they
answer, the two score 97.2 and 90.2. The sizing ranking therefore diverges from transcription's
(Figure~\ref{fig:curves}c) for two reasons: sizing knowledge is text-borne
and present at every Claude tier, and on the newest models a safety
policy, not capability, sets the score. Standardized per-class
measurement templates are the next refinement (\S\ref{sec:roadmap}). Because
the human sizes are public, Section~\ref{sec:contamination} treats answer-key
exposure explicitly.

\paragraph{Evaluation protocol.} The released harness evaluates any model
behind an API in a single invocation, and every request declares no tools,
so evaluations are closed book by construction. Outputs produced outside
the harness are scored from a standard submission format against the frozen
task list, and our own published runs pass through that same path, so
submitted and published results are graded by identical code. The scorer is
deterministic: identical inputs produce identical result tables across
processes and hash seeds, a property the parameter metric was corrected to
guarantee.

\paragraph{Baseline.} Table~\ref{tab:matrix} reports aggregate results, and
Figure~\ref{fig:curves} shows the per-task evidence behind them.

Six of seven models drop sharply between the small and medium tiers
(Figure~\ref{fig:tiers}, Appendix~\ref{app:tiers}); Haiku is already at floor on the small tier. The
failure among the remaining models is wiring, not reading:
a model that recovers a topology also reads its W/L annotations nearly
perfectly, so a single blended score would hide which skill is missing.
Tier difficulty is also not strictly monotonic: Opus and Gemini score higher
on small circuits than on trivial ones. The trivial tier is dominated by
passive networks and device models, whose vocabulary produces errors in models
that transcribe a five-transistor amplifier cleanly.

Failure mode separates the leading models as much as exact accuracy does
(Figure~\ref{fig:curves}a). Claude-fable-5 leads both exact matches and
structural F1. Behind it the frontier splits:
claude-opus-5 is more often fully correct than gpt-5.6-sol, but its failures
tend to be severe, while sol produces repairable near misses and holds the
second-best structural F1 despite a lower exact-match rate. The preferable model
therefore depends on whether a downstream system can repair a netlist.

The scorer itself required two corrections, reported because each would
have distorted a leaderboard. Parameter accuracy once scored against one
arbitrary isomorphism of each symmetric circuit; it now pairs devices
under the mapping with maximum parameter agreement among structurally valid
isomorphisms, selected deterministically rather than by hash-seed order. PDK resistor terminals were
once ordered pins, although the symbol draws no orientation and pin order
cannot be read from the image. The corrected scorer still rejects genuinely
misread connections, such as visually explicit bulk ties replaced by rail
defaults. Table~\ref{tab:matrix} also reports model-specific cutoff splits;
Section~\ref{sec:contamination} states what can and cannot be inferred from
them.

\begin{table}[h]
\centering\small
\caption{Model results on the fixed transcription set. Cutoffs are
vendor-reported; post $n$ counts tasks whose pinned revision postdates the
cutoff, ruling out that exact revision from the model's pretraining window.
Post/pre columns give exact isomorphism on each side. Sizing is the Task~2
score on a 100-point scale, with a proposal that provides no sizes scoring
zero; No ans.\ counts those proposals (refusals, one unparseable answer)
out of 17.}
\label{tab:matrix}
\begin{tabular}{lrrrrrrrr}
\toprule
Model & Cutoff & Iso & Struct & W/L & Post $n$ & Post / pre iso & Sizing & No ans. \\
\midrule
claude-fable-5         & 2026-01 & 56.1\% & 0.853 & 0.921 & 13 & 30.8\% / 62.3\% & 31.8 & 11 \\
claude-opus-5          & 2026-05 & 51.5\% & 0.711 & 0.929 & 0  & --- & 74.3 & 4 \\
gpt-5.6-sol            & 2026-02 & 42.4\% & 0.831 & 0.942 & 13 & 30.8\% / 45.3\% & 78.2 & 0 \\
gemini-3.1-pro-preview & 2025-01 & 39.4\% & 0.767 & 0.977 & 36 & 38.9\% / 40.0\% & 91.2 & 0 \\
claude-sonnet-5        & 2026-01 & 19.7\% & 0.604 & 1.000 & 13 & 7.7\% / 22.6\% & 85.4 & 0 \\
gpt-5                  & 2024-09 & 13.6\% & 0.512 & 0.988 & 38 & 15.8\% / 10.7\% & 79.0 & 0 \\
claude-haiku-4-5       & 2025-02 & 1.5\%  & 0.307 & 1.000 & 35 & floor & 78.3 & 1 \\
\bottomrule
\end{tabular}
\end{table}

\section{Measuring contamination}\label{sec:contamination}

Public GitHub repositories may enter model pretraining, so this benchmark
cannot assume independence. It instead provides circuit-level evidence for
probing potential exposure: pinned commit dates, blinded renders, release
canaries, and a closed-book protocol.

\paragraph{Commit-date analysis.} Each circuit retains its shuttle and pinned
commit date, which can be compared with a model's reported training cutoff.
Table~\ref{tab:matrix} performs this comparison per model. Claude-opus-5 has no
post-cutoff tasks in the fixed set. Across the two usable older-cutoff
comparisons, exact-match accuracy shows no consistent pre-cutoff advantage.
This descriptive comparison is not a contamination estimate: difficulty
differs across the date boundary, graph-hash groups are not separated across
it, and an earlier version of a design may predate the pinned revision.
The commit metadata supports the same analysis for future models, while new
shuttles expand the post-cutoff side.

\paragraph{Training-time and inference-time channels.} Commit dates probe
possible exposure in training data. A model evaluated with tool access has a second
channel: the upstream repositories are public, so a search-enabled agent can
retrieve an answer at inference time regardless of era. The protocol
separates the two. Baseline scores are closed-book: evaluation requests
declare no tools, so the model has no retrieval channel by construction.
Systems with search or repository access belong to a separate open-book
track and must be labeled as such. The blinded render removes author-selected
labels from the image. Reproducing an invisible internal name is evidence of
memorization or external retrieval, so the blinded condition also provides an
exposure diagnostic for submitted results.

\paragraph{Answer-key exposure differs by task.} For transcription the
artifact given to the model is an image that does not contain the netlist.
For sizing the upstream repository contains the human sizes, which is why
sizing is scored at perturbed operating points with regurgitation
diagnostics.

\paragraph{Canary.} Every released netlist carries the BIG-bench canary as an
ignored SPICE comment, and the release check verifies its placement. The canary
marks generated artifacts; it cannot protect upstream sources that were already
public. Commit-date splits address that channel, while blinded renders address
inference-time retrieval. Per-circuit memorization probes remain an archival
milestone.

\section{Limitations and roadmap}\label{sec:roadmap}

Most entries target Sky130~\cite{sky130}; broader coverage of GF180 and IHP
SG13G2, including SiGe, is planned from MPW-mirror, IHP, and ISHI-KAI sources.
Schematics with unresolved host paths are excluded from rendering-sensitive
splits. Circuit-class labels are keyword-derived and will be reclassified from
netlist graphs. Top-level designs without a parseable netlist remain outside
the graph-hash analysis.

Three milestones remain for the archival version. First, because author
testbenches expose heterogeneous measurements, we will add per-class templates
over the eligible set, starting with gain, gain-bandwidth product, and phase
margin for the amplifier class. This layer widens sizing coverage beyond the
leaf-subcircuit set. Second, the
per-circuit memorization probes described in
Section~\ref{sec:contamination}, joining the delivered blinded-render
comparison. Third, topology generation as a task. We do not
claim it in this version because two of its components are unsolved: a
uniform representation for the design brief, and automatic pin-matching
between a generated circuit and the author testbench. Both build on the
measurement layer; claiming the task before its scoring exists would
overstate readiness.

\section*{Acknowledgments}

We thank the Tiny Tapeout community and the designers whose open-source
circuits form this benchmark, and the AI4ChipDesign reviewers for their
feedback.

\section*{Ethics, licensing, and attribution}

All redistributed designs are Apache-2.0, verified from license texts
archived at the pinned commits; the released attribution file credits
every repository and its authors, and per-circuit manifests carry
source URL, commit, license, and extraction method. The Tiny Tapeout
index~\cite{tt} is CC0; the PDKs are Apache-2.0. Benchmark code and released
data use Apache-2.0, matching the source licenses. The code and data
are released at
\url{https://github.com/Li-Yueting/SGAnalog}, together with
a simulation browser that replays precomputed runs of the verified
pairs. Designer names
appear only as attribution of their public, licensed work. Per the
workshop's LLM policy: an LLM assistant was used in building the harvest
and evaluation harness and in drafting this manuscript; the authors
verified every released number against the committed tables, and the
release ships the consistency checker that enforces the match.

\bibliographystyle{plain}

\appendix

\section{Difficulty-tier breakdown of Task 1}\label{app:tiers}

\begin{figure}[h]
\centering
\includegraphics[width=0.62\textwidth]{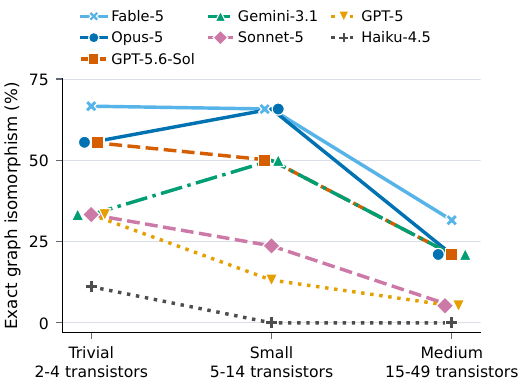}
\caption{Exact isomorphism by difficulty tier ($n=9, 38, 19$): five models
drop from small to medium, while Haiku is already at floor; Opus and Gemini
improve from trivial to small. Markers with identical scores are offset
slightly within each tier for visibility; only their vertical position
encodes accuracy.}
\label{fig:tiers}
\end{figure}

Figure~\ref{fig:tiers} plots exact isomorphism per difficulty tier for the
seven evaluated models, the per-tier view behind the tier claims in
Section~\ref{sec:tasks}.

\section{Full dataset inventory}\label{app:inventory}

Table~\ref{tab:inventory} expands the spine of Table~\ref{tab:funnel}
into the complete harvest, licensing, composition, and redundancy record.

\begin{table}[h]
\centering\small
\caption{The full dataset inventory. Left: the harvest funnel with triage
and licensing detail. Right: composition by role, transistor-count tier,
era, and device--net graph hash.}
\label{tab:inventory}
\begin{minipage}[t]{0.48\textwidth}
\centering
\begin{tabular}[t]{@{}lr@{}}
\toprule
Harvest funnel & $n$ \\
\midrule
Index projects (27 shuttles) & 4,572 \\
Entries with analog pins     & 237 \\
Unique repositories          & 195 \\
\quad ship xschem~\cite{xschem} schematics & 133 \\
\quad ship SPICE netlists only & 12 \\
\quad layout data only (unusable) & 48 \\
\quad fail to clone          & 2 \\
Usable repositories          & 145 \\
\quad license verified Apache-2.0 & 144 \\
\quad hold fetchable schematics & 129 \\
\quad contribute classified circuits & 125 \\
Schematics fetched at pinned commits & 1,448 \\
\bottomrule
\end{tabular}
\end{minipage}\hfill
\begin{minipage}[t]{0.50\textwidth}
\centering
\begin{tabular}[t]{@{}lr@{}}
\toprule
Composition & $n$ \\
\midrule
Circuits                     & 933 \\
\quad with a PDK transistor  & 809 \\
\quad top-level designs      & 425 \\
\quad sub-blocks             & 508 \\
Testbenches                  & 515 \\
\quad circuits driven by one & 350 \\
Top-level tiers: trivial (0--4 transistors) & 164 \\
\quad small (5--14) & 161 \\
\quad medium (15--49) & 78 \\
\quad large (50--149) & 20 \\
\quad very large (150+) & 2 \\
\quad non-trivial (5 or more) & 261 \\
Eras: 2023--24 / 2025 / 2026 & 194 / 143 / 88 \\
Graph-hashable top-level designs & 357 \\
\quad unique device--net graph hashes & 273 \\
\quad redundant designs & 84 \\
\quad duplicate groups / cross-repository & 43 / 32 \\
Fixed tasks: transcription / sizing & 66 / 17 \\
\bottomrule
\end{tabular}
\end{minipage}
\end{table}

\section{Simulation browser}\label{app:ui}

\begin{figure}[h]
\centering
\includegraphics[width=\textwidth]{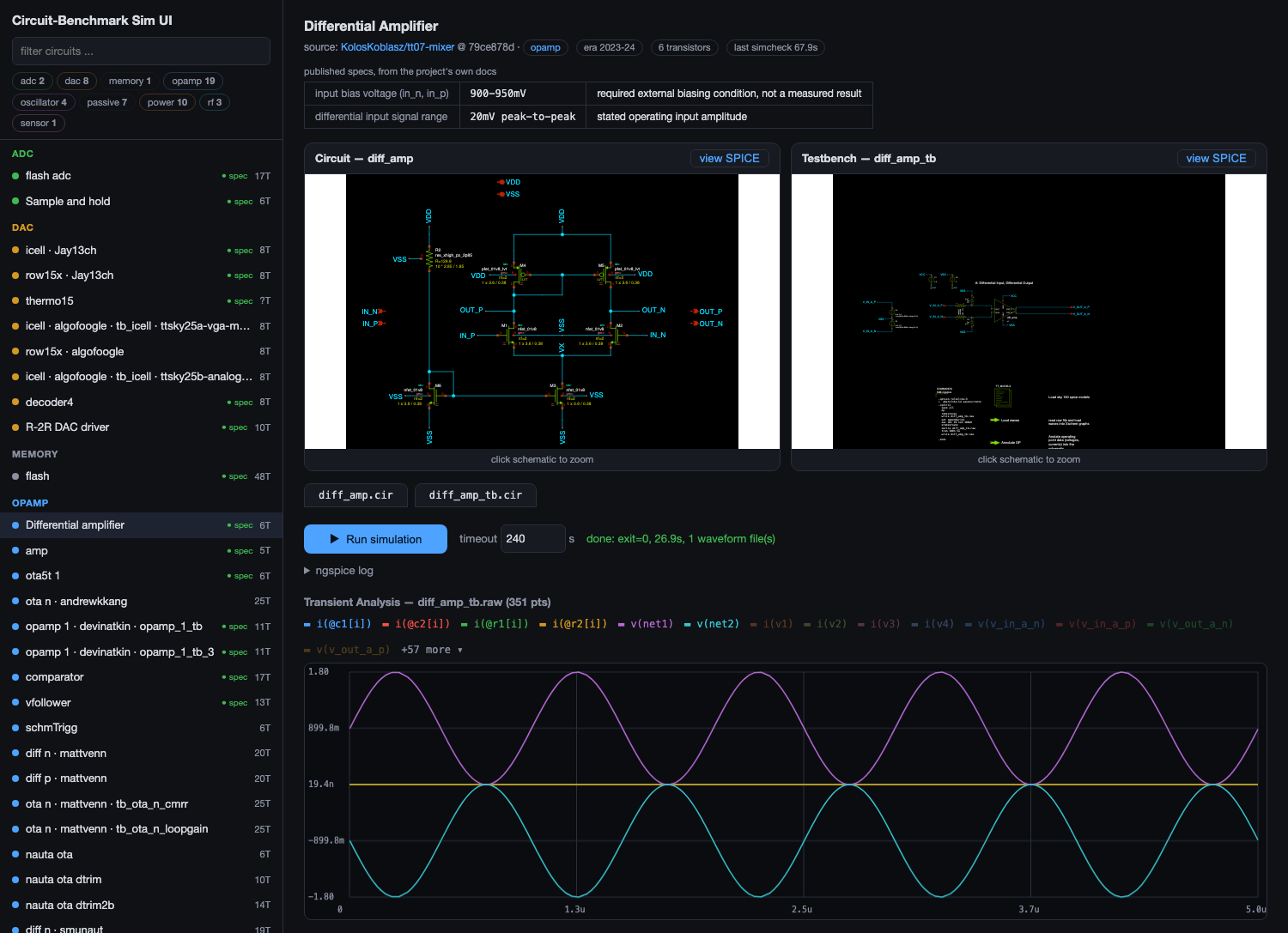}
\caption{The simulation browser on one verified pair, a differential
amplifier from the corpus: published specifications, circuit and
testbench schematics, and the transient waveforms produced by a run
inside the pinned container (exit 0 in 26.9 seconds; the two
complementary output sines around the input common mode).}
\label{fig:ui}
\end{figure}

The release includes a local browser interface over the verified
circuit--testbench pairs (Figure~\ref{fig:ui}). For each pair it shows
the circuit and testbench schematics, both extracted ngspice netlists,
the upstream repository at its pinned commit, and any specifications the
project's own documentation publishes. A run started from the page
executes through the same pinned-container entry point as the pipeline,
so its result is bit-identical to the batch simulatability check, and
the waveforms are parsed from the run's raw files. The sizing task set
of Section~\ref{sec:construction} is drawn from these verified pairs.
The released repository builds a static companion page with
precomputed runs of ten representative pairs, one per circuit class;
the live browser ships with the repository and requires only Docker.

\end{document}